\documentclass[10pt,twocolumn,letterpaper]{article}

\usepackage{cvpr}              % To produce the CAMERA-READY version
\usepackage{times}
\usepackage{epsfig}
\usepackage{graphicx}
\usepackage{amsmath}
\usepackage{amssymb}
\usepackage{algpseudocode}
\usepackage{multirow}
\usepackage{adjustbox}
\usepackage{booktabs}

\usepackage[pagebackref,breaklinks,colorlinks]{hyperref}

\usepackage[capitalize]{cleveref}
\crefname{section}{Sec.}{Secs.}
\Crefname{section}{Section}{Sections}
\Crefname{table}{Table}{Tables}
\crefname{table}{Tab.}{Tabs.}

\def\cvprPaperID{*****} % *** Enter the CVPR Paper ID here
\def\confName{CVPR}
\def\confYear{2024}

\begin{document}

\title{Consensus-Adaptive RANSAC}

\author{Luca Cavalli\\
ETH Z\"urich\\
{\tt\small luca.cavalli@inf.ethz.ch}
\and
Daniel Barath\\
ETH Z\"urich\\
{\tt\small danielbela.barath@inf.ethz.ch}
\and
Marc Pollefeys\\
ETH Z\"urich\\
{\tt\small marc.pollefeys@inf.ethz.ch}
\and
Viktor Larsson\\
Lund University\\
{\tt\small viktor.larsson@math.lth.se}
}
\maketitle
\newcommand{\todo}[1]{\textcolor{blue}{{[#1]}}}
\newcommand{\degrees}{$^\circ$}
\newcommand{\ac}[1]{#1}
\newcommand{\acs}[1]{#1}
\newcommand{\acl}[1]{#1}

\begin{abstract}
RANSAC and its variants are widely used for robust estimation, however, they commonly follow a greedy approach to finding the highest scoring model while ignoring other model hypotheses.
In contrast, Iteratively Reweighted Least Squares (IRLS) techniques gradually approach the model by iteratively updating the weight of each correspondence based on the residuals from previous iterations.
Inspired by these methods, we propose a new RANSAC framework that learns to explore the parameter space by considering the residuals seen so far via a novel attention layer.
The attention mechanism operates on a batch of point-to-model residuals, and updates a per-point estimation state to take into account the consensus found through a lightweight one-step transformer. 
This rich state then guides the minimal sampling between iterations as well as the model refinement.
We evaluate the proposed approach on essential and fundamental matrix estimation on a number of indoor and outdoor datasets. 
It outperforms state-of-the-art estimators by a significant margin adding only a small runtime overhead.
Moreover, we demonstrate good generalization properties of our trained model, indicating its effectiveness across different datasets and tasks.
The proposed attention mechanism and one-step transformer provide an adaptive behavior that enhances the performance of RANSAC, making it a more effective tool for robust estimation.
Code is available at https://github.com/cavalli1234/CA-RANSAC.
\end{abstract}

\begin{figure}[t]
\begin{center}
\includegraphics[width=\linewidth]{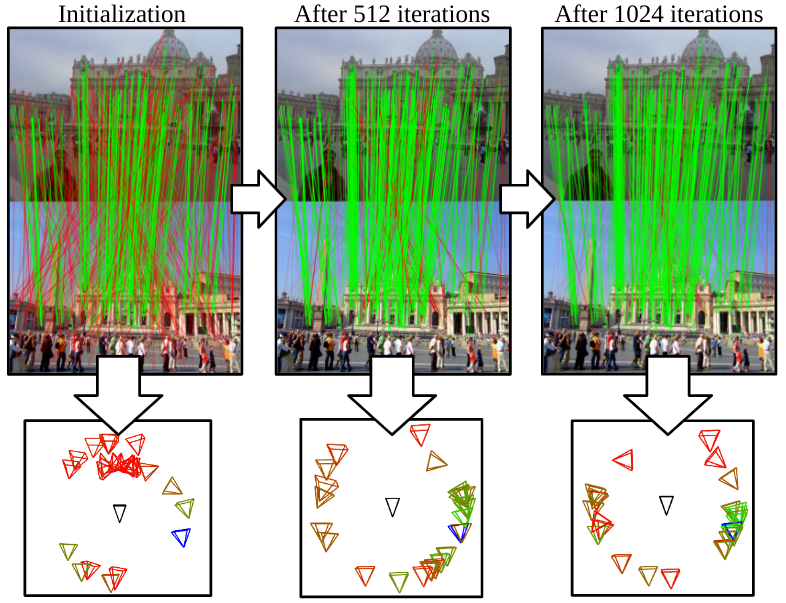}
\end{center}
   \caption{\textbf{Progression over iterations}: in the top row we show all correspondences (green for ground truth inliers, red for outliers) where each correspondence's opacity encodes our estimated inlier probability. Therefore, the visible correspondences have a high inlier likelihood estimate. The progression shows how CA-RANSAC incorporates the consensus found over the iterations to refine the initial rough inlier likelihood estimate. This results in much less visible outliers from left to right. On the bottom row we show camera models estimated from minimal samples taken from our proposed sampler, colored according to their MSAC score (green is better). Over iterations, our sampling scheme adjusts dynamically to become more and more focused around the correct model, shown in blue.}
\label{fig:teaser}
\end{figure}

%%%%%%%%% BODY TEXT
\section{Introduction}

Robust model estimation is a ubiquitous problem in computer vision, crucial for a wide range of tasks such as visual localization, Structure-from-Motion, Simultaneous Localization and Mapping (SLAM), and object pose estimation.
When having noisy and outlier-contaminated data, the established approach to achieving accurate and robust model estimation is the Random Sample Consensus (RANSAC) algorithm~\cite{fischler1981random}. 
RANSAC iterates selecting a minimal sample of data points, estimating the model parameters, and calculating the model quality via its support, \ie the number of measurements consistent with it.
Despite its popularity, RANSAC has several limitations, such as sensitivity to the inlier-outlier threshold parameter, that has inspired significant efforts throughout the years to improve it. 
Such advancements include novel sampling ~\cite{chum2005matching,torr2002napsac,cavalli2022nefsac,brachmann2019neural}, model scoring~\cite{barath2019magsac,barath2019magsacpp,barath2022learning} and
refinement algorithms~\cite{chum2003locally,lebeda2012fixing,barath2022learning}, and degeneracy checks~\cite{chum2005two,cavalli2022nefsac} to improve the accuracy, run-time and robustness to outliers.
More recently, methods focus on learning specific aspects (\eg, scoring or sampling functions) of the RANSAC framework~\cite{brachmann2019neural,barath2022learning,cavalli2022nefsac} or to make the estimation differentiable end-to-end~\cite{brachmann2017dsac,wei2022fully}.

While RANSAC and its classical or learned variants are widely used for robust estimation, one of their limitations is their greedy approach to finding the model with the highest score while ignoring sub-optimal model hypotheses along the way. 
In contrast, Iteratively Reweighted Least Squares (IRLS) techniques~\cite{rousseeuw1984least,ranftl2018deep} gradually approach the model by iteratively updating the weight of each correspondence based on the residuals from previous iterations.

Inspired by these methods, we propose CA-RANSAC, a new RANSAC framework that learns to explore the parameter space better by considering the residuals seen so far via a novel attention layer.
The attention mechanism operates on a batch of point-to-model residuals, and updates a per-point estimation state to take into account the consensus found through a lightweight one-step linear transformer. 
This rich state then guides the minimal sampling between iterations as well as the non-linear model refinement. 
This approach is trained end-to-end supervising directly on the quality of the estimated model and estimated inlier likelihoods.
We evaluate the proposed approach on essential and fundamental matrix estimation on a number of indoor and outdoor datasets. 
We find that CA-RANSAC outperforms state-of-the-art estimators by a significant margin adding only a small runtime overhead.
Moreover, we demonstrate good generalization properties of our trained model, indicating its effectiveness across different datasets and tasks.
The proposed attention mechanism and one-step linear transformer provide an adaptive behavior that enhances the performance of RANSAC, making it a more effective tool for robust estimation.

Unlike existing RANSAC variants that use pre-estimated probabilities, ours are updated and refined during RANSAC as a consequence of their implied consensus.
This leads to progressively sampling the best data points that help in finding and refining the sought model, as shown in Figure~\ref{fig:teaser}.
In summary, this paper brings the following contributions:

\begin{itemize}
\item It proposes a novel RANSAC framework that dynamically leverages the consensus found in previous iterations to enhance sampling and model refinement.

\item It introduces a simple and novel attention scheme that effectively encodes consensus information.

\item It compares the proposed method under various conditions (E/F matrix, SuperGlue/HardNet matches, Indoor/Outdoor) with the State of the Art, showing substantial improvements and good generalization.
\end{itemize}

\begin{figure*}
\begin{center}
\includegraphics[width=\linewidth]{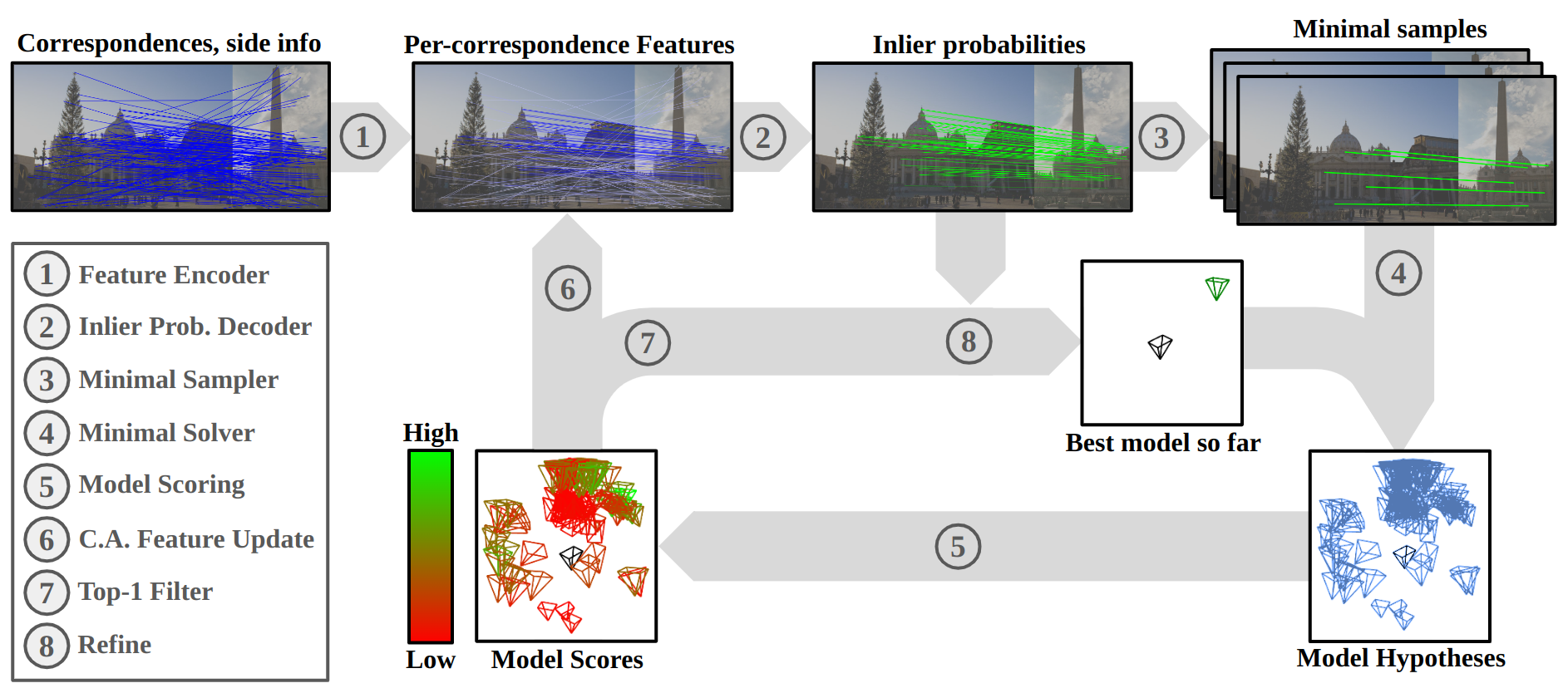}
\end{center}
   \caption{\textbf{CA-RANSAC architecture}. Given image correspondences and side information as input we \textbf{(1)} encode them to per-correspondence latent states, \textbf{(2)} decode the state to estimate the inlier probabilities, \textbf{(3)} take a batch of minimal samples from correspondences with high-enough estimated inlier probability and \textbf{(4)} solve for model hypotheses from minimal samples including the best model found so far. We then \textbf{(5)} score models with MSAC scoring, and \textbf{(6)} use the score matrix to update our per-correspondence latent states with consensus-aware attention. \textbf{(7)} The best model hypothesis of the batch is selected for \textbf{(8)} global refinement weighted by the inlier probabilities iteratively re-estimated from the updated state. Steps (2) to (8) are iterated multiple times.}
\label{fig:arch}
\end{figure*}

\section{Related Work}

Robust model estimation deals with estimating a model given a set of outlier-contaminated data points, each representing some constraints to the sought model. 
%This problem is so ubiquitous in computer vision that enormous research effort went into designing and improving techniques to address robust estimation in a variety of scenarios. 
%Despite being introduced in the early 80's,
%The by far most common approaches are based on 
%
%The field is currently dominated by hypothesize-and-verify methods based on the original RANSAC~\cite{fischler1981random} introduced in the early 80's. 
The current paradigm is to use sampling based methods such as  RANSAC~\cite{fischler1981random}, which
is one of the most successful and popular techniques for robust model estimation from image correspondences. Given its core importance, extensive research has gone into improving its robustness, accuracy, and runtime, which we review in this section.
%Still, nowadays, RANSAC~\cite{fischler1981random} is one of the most successful and popular techniques applied to robust model estimation from image correspondences. Given its core importance, extensive research has gone into improving its robustness, accuracy, and runtime, which we review in this section.

Several approaches have been proposed to replace the uniform sampling in RANSAC and increase the probability of finding an all-inlier sample and a good model early.
PROSAC~\cite{chum2005matching} assumes that the data points have been pre-ordered by a prior probability and draws the most likely minimal samples first, progressively transitioning to uniform sampling until a good model is found. 
This is performed by sorting the tentative point correspondences according to some quality metric, \eg, the second nearest neighbor (SNN) ratio, which then serves as a proxy for the inlier likelihood.
Other approaches explore conditional sampling to increase the success rate, such as sampling neighboring correspondences~\cite{torr2002napsac} or from clusters~\cite{ni2009groupsac}.
Fan et al.~\cite{fan2009adaptive} propose a heuristics to adaptively bias sampling towards successful models. While close in motivation to our work, our framework is more general and does not rely on handcrafted heuristics.

More recently, there has been growing interest in learning sampling distributions to accelerate RANSAC. 
Neural Guided RANSAC~\cite{brachmann2019neural} learns a categorical sampling distribution through an approximate expected gradient. 
At the same time, $\nabla$-RANSAC~\cite{wei2022fully} proposes a fully differentiable RANSAC that learns a categorical sampling distribution using the Gumbel re-parameterization trick, enabling end-to-end learning. 
These approaches aim to increase the chance of finding an outlier-free minimal sample early by prioritizing samples more likely to be inliers.

% A substantial acceleration in modern RANSAC implementation comes from~\cite{chum2008optimal}, which optimizes each iteration by applying a modified sequential probability ratio test (SPRT) to preempt unpromising model verification.

Several works have been proposed to improve the quality calculation in RANSAC beyond uniform sampling. 
Moisan et al.~\cite{moisan2012automatic} and Stewart et al.~\cite{stewart1995minpran} propose alternative distributions to model the noise in the residuals, while Bayesian RANSAC~\cite{torr2002bayesian} and MLESAC~\cite{torr2000mlesac} use Bayesian inference and maximum likelihood estimation to estimate the model parameters and inliers, respectively. 
DSAC~\cite{brachmann2017dsac} uses probabilistic model selection to enable learning the model scoring together with the features.
Additionally, MAGSAC++~\cite{barath2019magsac,barath2019magsacpp,barath2021marginalizing} marginalizes over a range of noise scales to minimize sensitivity to the inlier-outlier threshold.
Other approaches aim to integrate local optimization, degeneracy checks, and pre-emptive model verification into RANSAC to improve its accuracy and robustness. 
LO-RANSAC~\cite{chum2003locally} performs local optimization of the best models found during the model search, greatly improving the accuracy.
The original local optimization has been further improved in \cite{lebeda2012fixing} and with graph-cut segmentation of outliers in \cite{barath2018graph}. 
DEGENSAC~\cite{chum2005two} recognizes that degenerate models can lead to high consensus while being inconsistent with the scene geometry, and checks for planar degeneracy in minimal samples. 
QDEGSAC~\cite{frahm2006ransac} deals with quasi-degenerate models by actively searching for the missing constraints within the outlier set.
USAC~\cite{raguram2013usac} and VSAC~\cite{ivashechkin2021vsac} integrate progressive sampling, local optimization, SPRT, and other improvements into a single framework.

Other approaches aim to prune outliers before RANSAC is applied. 
Bian et al.~\cite{bian2017gms} and Cavalli et al.~\cite{cavalli2020handcrafted} use explicit local verification of neighboring correspondence statistics or local affine models, respectively. 
Yi et al.~\cite{yi2018learning} apply batch normalization as context within a PointNet~\cite{qi2017pointnet} to learn outlier rejection. 
\cite{sun2020acne} introduce an attention mechanism to make the context normalization robust. 
OANet~\cite{zhang2019learning} proposes order-aware pooling of point features to learn a soft clustering. 
CLNet~\cite{zhao2021progressive} proposes architectural improvements to progressively prune outliers, propagating gradients from models to points through the weighted eight-point algorithm. 
MS2DG-Net~\cite{dai2022ms2dg} builds a dynamic sparse semantic graph to better mine spatial correlations between inliers. 
Some outlier filters also produce model estimates to guide the outlier segmentation through consensus, as proposed by \cite{zhao2021progressive,dai2022ms2dg}. 
However, it has been shown that a RANSAC re-estimation on the filtered inliers is always beneficial~\cite{cvpr2020ransactutorial}.

We propose a novel use of correspondence consensus within RANSAC for gradually refining inlier probabilities. 
Unlike existing RANSAC variants that use pre-estimated probabilities, ours are updated and refined \textit{during} RANSAC as a consequence of their implied consensus.
This leads to progressively sampling the best data points that help in 
finding and refining the sought model, as shown in Figure~\ref{fig:teaser}. 

\section{Consensus-Adaptive RANSAC}

Let us assume that we are given a set of data points $\mathcal{D} = \{ x \in \mathbb{R}^p \}$, consisting of outliers and noisy inliers. 
The objective is to find a model $M \in \mathbb{R}^q$ that minimizes a fitting cost $\mathcal{C} = \sum_{x \in \mathcal{D}} \mathcal{L}(\mathcal{R}(M, x))$ where $\mathcal{R} : \mathbb{R}^q \times \mathbb{R}^p \mapsto \mathbb{R}$ is the error of a data point $x$ with respect to a model $M$, and $\mathcal{L} : \mathbb{R} \mapsto \mathbb{R}$ is a robust loss.
The role of $\mathcal{L}$ is to minimize the influence of outlier data points. Consequently, it assigns small or zero gradients for large errors.

RANSAC~\cite{fischler1981random} explores the model space in a discrete manner by fitting a set of models $M_i^{min}$ to a random subset of the data $\mathcal{D}_i^{min} \subset \mathcal{D}$ in each iteration.
To minimize the risk of outlier-contamination, the subsets $\mathcal{D}_i^{min}$ are chosen to be as small as possible, \ie, they are minimal samples.
%RANSAC~\cite{fischler1981random} explores the model space in a discrete manner by iteratively sampling minimal sets of data points $D_i^{min}$ such that they consist of the minimum number of points $x$ that can fit exactly a finite set of models $M_i^{min}$. 
%Using minimal sets minimizes the chances to pick an outlier, so that a non-robust estimator can be used to obtain a rough estimate of the underlying model.
Each model $M^{min}_i$ is then checked against every data point, computing a set of residuals $R_i = \{\mathcal{R}(M_i^{min}, x) \; | \; x \in \mathcal{D}\}$.
For each model, the residuals are used to derive a consensus score $c_i \in \mathbb{R}$, and the best scoring model is kept.
%from which a consensus score $c_i$ is derived.
%This score allows us to compare different models, and the best one is kept.
%This score is used to select the model with the highest consensus. 
The consensus score $c_i$ can for example be taken as the size of the consensus set 
$c_i = \#\{r \in R_i \; | \; r < T\}$ for some threshold $T\in\mathbb{R}_+$, or the average truncated model agreement,
$c_i = \frac{1}{|\mathcal{D}|}\sum_{r \in R_i} \min(r,T)$.
%The consensus set is calculated as the number of inliers $c_i = \#\{r \in R_i \; | \; r < T\}$ where an inlier has an error smaller than a manually set inlier threshold $T$. 
For the best accuracy, the final model is then optimized using inlier measurements, directly minimizing the fitting cost $\mathcal{C}$.

%In this paper, we generalize the consensus estimation in RANSAC to be amenable to a transformer-like architecture. Our generalization enables learning to account for the model consensus found between data points, which we show to improve sampling efficiency and model refinement.

In this paper, we revisit the RANSAC framework by rethinking the consensus score and using the model affinities as a form of attention between correspondences.
We decorate each correspondence with a latent state $f_i \in \mathbb{R}^c$, from which we learn to predict inlier probabilities, used for sampling and model refinement.
Over iterations, the latent estimation state $\mathcal{F} = \{ f_i \in \mathbb{R}^c \}_{i=1..|\mathcal{D}|}$ is repeatedly updated by a lightweight one-step transformer using the residual-driven attention to condition the state on the consensus observed so far, thus achieving a consensus-adaptive robust estimation algorithm.

%The residual-driven attention is input to a light-weight transformer that updates the per-correspondence state  $\mathcal{F} = \{ f_i \in \mathbb{R}^c \}_{i=1..n}$, achieving a consensus-adaptive robust estimation algorithm.

%In this paper, we first generalize the consensus scoring in RANSAC to an attention mechanism among correspondences. 
%Then, we use this attention mechanism in a lightweight one-step transformer to update a per-correspondence estimation state $\mathcal{F} = \{ f_i \in \mathbb{R}^c \}_{i=1..n}$, where each $f_i$ is a $c$-dimensional feature vector associated directly to each data point $x_i$. 
%Finally, we use the consensus-aware state $\mathcal{F}$ to condition the sampling and refinement steps in RANSAC, achieving consensus-adaptive sampling and refinement.

\subsection{Consensus-based Attention}
\label{subsec:consensus_attention}
Let $r_{ij} = \mathcal{R}(M_j, x_i)$ be the residual of model $M_j$ with data point $x_i$. Let $\mathcal{S}: \mathbb{R} \mapsto [0, 1]$ be a scoring function that maps residuals $r_{ij}$ to scores $s_{ij}$, where a higher score is associated with a stronger consensus of point $x_i$ with model $M_j$. 
While function $\mathcal{S}$ itself could be trainable and could be conditioned on the latent states, in our experiments we did not observe significant improvements compared to the MSAC definition $\mathcal{S}(r) = 1 - \min(r, T)/T$ for a sensible choice of threshold $T$, thus, we use the latter.

Let $S \in [0, 1]^{n \times m}$ be the score matrix of $n$ data points over $m$ model hypotheses, and $C_j = \sum_{i=1}^n s_{ij}$ be the total consensus gathered by model $j$. We define our novel normalized attention matrix $A \in [0, 1]^{n \times n}$ as follows:
\begin{equation} \label{eq:attention_matrix}
A =  \frac{SS^T}{\sum_{j=1}^m C_j}.
\end{equation}

Matrix $A$ is such that $A_{ik} \propto \sum_{j=1}^m s_{ij}s_{kj}$ measures the correlation of the agreement of data points $x_i$
 and $x_k$ on the available models, generalizing the graph edge scores in~\cite{zhao2021progressive}. One major property of $A$ is that, differently from traditional attention mechanisms that require row-normalization for stable learning, it does not require any normalization. In fact, for a point $\hat{i}$ the following holds:
 \begin{equation}
 \sum_{k=1}^n A_{\hat{i}k} = \sum_{k=1}^n \frac{\sum_{j=1}^m s_{\hat{i}j}s_{kj}}{\sum_{l=1}^m C_l} =  \sum_{j=1}^m s_{\hat{i}j}\frac{C_j}{\sum_{l=1}^m C_l}.
 \end{equation}
 Therefore, the row sum directly quantifies the consensus gathered by the corresponding data point, weighted by each model's relative consensus. This makes an excellent signal to quantify the overall consensus and attribute it to other data points. Moreover, the bounds for row sums are trivially obtained for a point with zero or perfect consensus with all models ($s_{\hat{i}j} = 0$ or $1$ for every $j$), giving the desired bound of $[0, 1]$.
 Note that this is in contrast with traditional normalization techniques which enforce that each row sums exactly to one, thus canceling out the information about the total consensus of the corresponding data point.

 This key component in CA-RANSAC enables the following state update procedure:
 \begin{equation} \label{eq:feat_update}
     \mathcal{F} \leftarrow \text{MLP}([\mathcal{F}, \text{MLP}(A \cdot \text{MLP}(\mathcal{F}))]),
 \end{equation}
where each MLP is a separate Multi Layer Perceptron sharing weights across all data points, $\cdot$ is matrix multiplication, and $[ \, \cdot \, ]$ indicates concatenation.

We observed experimentally that more expressive state update schemes, in multiple steps and with data-dependent attention gated by matrix $A$ as in standard transformers lead to only minor improvements, with a great cost in terms of additional computational complexity.

\subsection{Algorithmic Structure}
\label{subsec:algo}

%In RANSAC, all data points are treated equally. 
%However, external knowledge about each data point is often available, which can be used for more efficient and effective processing. 
%A traditional example in fundamental matrix estimation is the second nearest neighbor descriptor distance, frequently used to reject correspondences. 
%Recent works~\cite{zhao2021progressive,dai2022ms2dg,zhang2019learning} show that, even without extra information, a network can learn to distinguish patterns among image correspondences for outlier filtering. 
%With the proposed Consensus-Adaptive RANSAC, we wish to integrate such knowledge within the procedure, condition it on the found consensus, and make it propagate it to components in RANSAC, such as sampling and model refinement.

Given the update mechanism outlined in Section~\ref{subsec:consensus_attention}, the latent state $\mathcal{F}$ can be conditioned on the consensus found in previous iterations. 
In this Section, we describe the proposed CA-RANSAC algorithm that takes advantage of the rich information stored in the consensus-aware state $\mathcal{F}$. 

Let $n = |\mathcal{D}|$ be the number of data points available, and $\mathcal{E} = \{ e_i \in \mathbb{R}^d \}_{i=1..n}$ be the external information, \eg SNN ratio, available for each point. CA-RANSAC can be described as the following algorithm, also outlined in Fig.~\ref{fig:arch}:

\begin{algorithmic}
\Require $\mathcal{D}$, $\mathcal{E}$ -- data points and external information
\Ensure $\hat{M}^*$ -- final model params., $\mathcal{I}$ -- inlier probabilities 
\State $\hat{M}^* \leftarrow \textbf{0}$
\State $\mathcal{F} \leftarrow \text{\textsc{InitState}}_{\mathbf{\gamma}}(\mathcal{D}, \mathcal{E})$
\State $\mathcal{I} \leftarrow \text{\textsc{InlierDecoder}}_{\mathbf{\theta}}(\mathcal{F})$
\While{$\neg$\textsc{Terminate}()}
\State $\mathcal{D}^{min} \leftarrow \text{\textsc{BatchedSampler}}(\mathcal{D}, \mathcal{I})$ 
\State $\hat{M} \leftarrow \text{\textsc{EstimateModel}}(\mathcal{D}^{min}) \cup \{\hat{M}^*\}$
%\State $S \leftarrow \text{MSAC}(\hat{M}, \mathcal{D})$
%\State $\hat{M}^{min} \leftarrow \text{top-k}(\hat{M}^{min}, S)$
%\State $\hat{M} \leftarrow \text{RefineTruncated}(\hat{M}^{min}, \mathcal{D})$
\State $S \leftarrow \text{\textsc{ScoreModels}}(\hat{M}, \mathcal{D})$
\State $A \leftarrow \text{\textsc{ConsensusAttention}}(S)$ \Comment{Eq.~\ref{eq:attention_matrix}}
\State $\mathcal{F} \leftarrow \text{\textsc{StateTransformer}}_{\mathbf{\omega}}(\mathcal{F}, A)$ \Comment{Eq.~\ref{eq:feat_update}}
\State $\mathcal{I} \leftarrow \text{\textsc{InlierDecoder}}_{\mathbf{\theta}}(\mathcal{F})$
\State $\hat{M}^* \leftarrow \text{\textsc{top-one}}(\hat{M}, S)$
\State $\hat{M}^* \leftarrow \text{\textsc{Refine}}_\alpha(\hat{M}^*, \mathcal{D}, \mathcal{I})$
\EndWhile
\end{algorithmic}

We implement both \textsc{InitState}\textsubscript{$\gamma$} and \textsc{InlierDecoder}\textsubscript{$\theta$} as MLPs with parameters $\gamma$ and $\theta$, shared across every correspondence. 
\textsc{InitState}\textsubscript{$\gamma$} is used to initialize the latent state from additional side information (usually, the SNN ratio from the matching), and \textsc{InlierDecoder}\textsubscript{$\theta$} decodes the state to estimate the inlier probabilities. Note that \textsc{InitState}\textsubscript{$\gamma$} could be designed to be more expressive, taking inspiration from outlier rejection networks~\cite{yi2018learning,zhang2019learning,zhao2021progressive,dai2022ms2dg}, but we show that even a simple design is very effective when coupled with the signal from consensus.

The \textsc{BatchedSampler} draws a batch of $m$ minimal samples uniformly at random from the pool of points $\{x_i \in \mathcal{D}~|~p_i > T\}$ where $p_i \in \mathcal{I}$ is the estimated probability that point $i$ is an inlier. The threshold $T=0.4$ is chosen to guarantee a $90\%$ probability to sample a full-inlier sample in $N_{iter}=256$ trials with an inlier ratio of $T$. When the pool is smaller than $M_{min}=15$ samples, then the $M_{min}$ best scoring correspondences are taken as a pool regardless of their absolute score. Note that a standard PROSAC~\cite{chum2005matching} cannot be trivially applied in our case since the inlier estimates are bound to change over iterations.
\textsc{ScoreModels} returns the scoring matrix $S$ as outlined in Section~\ref{subsec:consensus_attention}, using the squared Sampson distance as residual function $\mathcal{R}$.
\textsc{ConsensusAttention} builds the attention matrix $A$ following Equation~\ref{eq:attention_matrix}, and the \textsc{StateTransformer}\textsubscript{$\omega$} uses it to update the state $\mathcal{F}$ according to Equation~\ref{eq:feat_update}. 
The \textsc{Refine}\textsubscript{$\alpha$} step performs robust nonlinear refinement of the best model found so far, weighting data points proportionally to a learned power $\alpha$ of the inlier probabilities $\mathcal{I}$. We use a PyTorch implementation of the Levenberg-Marquardt algorithm~\cite{more1978levenberg} and propagate gradients over the last refinement step to train the parameter $\alpha$ initialized to 1.

In principle, the standard RANSAC termination criterion can be used in CA-RANSAC. 
However, we run CA-RANSAC always for four iterations, each batching 256 samples, to be able to compare different robust estimators on the same number of iterations as described in Section~\ref{sec:experiments}. We report further implementation details in Appendix~\ref{sec:impl_details}.

%A detailed description of each step of our implementation can be found in the supplementary material.
\subsection{Supervision}

\begin{table}[t]
\begin{center}
        \caption{Ablation study on CA-RANSAC. We compare the proposed method in full (Full), ours without consensus-aware feature update (-C), ours without adaptive sampling (-S), ours without adaptive refinement (-R) and our closest traditional baseline (LM-LO). We run the comparison on HardNet mutual nearest neighbor matches on DoG detections (DoG+HN) on PhotoTourism (PT) and on SuperGlue matches (SP+SG) on ScanNet (SN).}
\label{tab:ablations}
\resizebox{\linewidth}{!}{
\begin{tabular}{lccccc} 
\toprule
        & Matcher & Dataset & AUC5 & AUC1 & MAP20 \\
\midrule
        LM-LO & \multirow{5}{*}{DoG+HN} & \multirow{5}{*}{PT} & 63.9 & 34.0 & 91.8\\
-C &  &  &  55.8 & 28.8 & 88.8 \\
-S &  &  &  61.5 & 33.3 & 82.7 \\ 
-R & &  & \bf 69.3 & 36.5 & 96.1 \\
Full &  &  & 69.1 & \bf 37.4 & \bf 96.3 \\
\midrule
        LM-LO & \multirow{3}{*}{SP+SG} & \multirow{3}{*}{SN} &14.3 & 0.3 & 72.9 \\
-R &  &  & 15.1 &  0.3 & 74.5 \\
Full &  &  & \bf 19.5 & \bf 0.8 & \bf 77.7 \\
\bottomrule
\end{tabular}
}
\end{center}
\end{table}

We train CA-RANSAC end-to-end using the ground truth model $\mathcal{M}^*$. 
In particular, we address the problems of essential and fundamental matrix estimation and train using the ground truth relative pose and calibration. 
We label every correspondence an inlier if its Sampson distance is smaller than 1 pixel w.r.t.\ the ground truth model. 
Let $p^q_i$ be the predicted inlier probability of data point $i$ at iteration $q$, let $l_i$ be its binary inlier label, and let $\mathcal{X}$ be the binary cross-entropy loss. 
Then we define the inlier classification loss as follows:
\begin{equation}
    \mathcal{L}^{q}_{inl} = \frac{\sum_{i=1}^n\mathcal{X}(p^q_i, l_i)}{n},
\end{equation}
where $n$ is the total number of data points in an image pair. This signal calibrates the inlier probabilities and trains the state initialization and update to make inliers separable from outliers. However, $\mathcal{L}_{inl}$ is not sufficient to supervise the refinement parameter $\alpha$ effectively. We, therefore, introduce a loss on the estimated model:
\begin{equation}
    \mathcal{L}^{q}_{pose} = \mathcal{E}(\hat{M^q}^*, M^*),
\end{equation}
where $M^*$ is the ground truth model, $\hat{M^q}^*$ is the model estimate until iteration $q$, and $\mathcal{E}$ is a model error measure.
In practice, we use the maximum between translation and rotation errors for both essential matrix estimation and fundamental matrix estimation, using the ground truth calibration in the latter case. 
We clamp $\mathcal{L}^q_{pose}$ at 30$^\circ$ since the supervision on local refinement is only informative when the estimated solution is close to the correct solution.

We weigh $\mathcal{L}_{pose}$ and $\mathcal{L}_{inl}$ and aggregate the total loss of all $Q$ iterations with exponential terms giving more weight to the last iteration batch as follows:
\begin{equation}
\mathcal{L} = \sum_{q=1}^Q (1-\epsilon)^{Q-q} ( \mathcal{L}^q_{inl} + \lambda \mathcal{L}^q_{pose})
\end{equation}
We set $\epsilon = 0.1$ and $\lambda = 1/60$ to scale $\mathcal{L}_{pose}$ in degrees to a similar order of magnitude to $\mathcal{L}_{inl}$.

\section{Experimental Results}
\label{sec:experiments}

We validate CA-RANSAC on robust essential and fundamental matrix estimation from a set of outlier-contaminated correspondences in outdoor and indoor settings.
We measure the maximum of the rotation and translation errors in degrees for every image pair. 
We employ the same procedure for fundamental matrix estimation, where the ground truth calibration is used to upgrade the final estimated fundamental matrix to an essential matrix for error calculation. 
From this error signal, we distill several metrics: we report average pose error (Avg) and Mean Average Precision under 20$^\circ$ (MAP20) as proxies for pose estimation robustness, median pose error (Med) and Area Under the Curve below 1$^\circ$ (AUC1) as a proxy for pose refinement accuracy, and Area Under the Curve below 5$^\circ$ (AUC5).

We run CA-RANSAC for a fixed number of 4 batched iterations, where each batch takes 256 samples, for a total of 1024 equivalent RANSAC iterations and 4 rounds of state updates. Comparably, we run every other RANSAC baseline for exactly 1024 iterations, unless specified differently. We refer to Appendix~\ref{sec:processing_time} for an analysis of the processing time of CA-RANSAC in this setting.

\subsection{Datasets}

\begin{table}[t]
\begin{center}
        \caption{\textbf{Essential matrix} estimation on \textbf{SuperGlue} matches from 9900 images of the \textbf{PhotoTourism} validation set.
        We measure pose error as the maximum between rotational and translational error in degrees. We report average error (Avg), median error (Med), Mean Average Precision under 20 degrees (\ac{MAP}@20\degrees), Area under the Curve under 5 degrees (\ac{AUC}@5\degrees) and under 1 degree (\ac{AUC}@1\degrees).
}
\label{tab:photo_spsg_ess}
\resizebox{\linewidth}{!}{
\begin{tabular}{lccccc}
\toprule
        & AUC5 & AUC1 & MAP20 & Med & Avg \\
        \midrule
        CA-RANSAC & \bf 75.1 & \bf 38.1 & \bf 98.8 & \bf 0.627 & \bf 2.25\\
        \ac{USAC} & 69.9 & 32.7 & 98.2 & 0.806 & 3.06\\
        LM-LO & 71.7 & 34.7 & 98.6 & 0.733 & 2.57\\
        \ac{MAGSAC}++ & 70.6 & 34.2 & 96.5 & 0.740 & 4.20\\
\bottomrule
\end{tabular}
}
\end{center}
\end{table}

\begin{table}[t]
\begin{center}
        \caption{\textbf{Essential matrix} estimation on \textbf{HardNet} matches from 9900 images of the \textbf{PhotoTourism} validation set. Every baseline uses the \ac{SNN} filter for the best performance, expect for \ac{CA-RANSAC} which learns to make use of the \ac{SNN} ratio as side information. LM-LO (100k) runs for 100.000 iterations and LM-LO (10k) runs for 10.000 iterations.
%We measure pose error as the maximum between rotational and translational error in degrees. We report average error (Avg), median error (Med), Mean Average Precision under 20 degrees (\ac{MAP}@20\degrees), Area under the Curve under 5 degrees (\ac{AUC}@5\degrees) and under 1 degree (\ac{AUC}@1\degrees).
}
\label{tab:photo_dog_ess}
\resizebox{\linewidth}{!}{
\begin{tabular}{lccccc}
\toprule
        & AUC5 & AUC1 & MAP20 & Med & Avg \\
\midrule
        \ac{CA-RANSAC} & \bf 69.1 & \bf 37.4 & \bf 96.3 & \bf 0.691 & \bf 4.75\\
        \ac{USAC} & 60.7 & 31.1 & 90.6 & 1.020 & 9.73\\
        LM-LO & 63.9 & 34.0 & 91.8 & 0.845 & 8.84\\
        LM-LO (10k) & 66.4 & 35.6 & 93.2 & 0.764 & 7.41\\
        LM-LO (100k) & 66.6 & 35.7 & 93.6 & 0.761 & 7.06\\
        \ac{MAGSAC}++ & 57.2 & 27.7 & 88.3 & 1.240 & 11.5\\
        GC-RANSAC & 63.1 & 33.2 & 91.4 & 0.873 & 8.85 \\
        VSAC & 61.0 & 31.0 & 92.3 & 1.01 & 8.49 \\
\bottomrule
\end{tabular}
}
\end{center}
\end{table}

\begin{table}[t]
\begin{center}
        \caption{\textbf{Essential matrix} estimation on \textbf{HardNet} matches \textbf{filtered} with \ac{CLNet}~\cite{zhao2021progressive} or \ac{OANet}~\cite{zhang2019learning} from 9900 images of the \textbf{PhotoTourism} validation set. Every baseline uses \ac{CLNet} or \ac{OANet} as filter for the best performance, expect for \ac{CA-RANSAC} which learns to make use of pre-trained \ac{CLNet} features as side information.
% We measure pose error as the maximum between rotational and translational error in degrees. We report average error (Avg), median error (Med), Mean Average Precision under 20 degrees (\ac{MAP}@20\degrees), Area under the Curve under 5 degrees (\ac{AUC}@5\degrees) and under 1 degree (\ac{AUC}@1\degrees).
}
\label{tab:photo_dog_filt_ess}
\resizebox{\linewidth}{!}{
\begin{tabular}{lccccc}
\toprule
        & AUC5 & AUC1 & MAP20 & Med & Avg \\
\midrule
        \ac{CLNet} & 51.5 & 18.5 & 92.0 & 1.884 & 7.39\\
        + \ac{CA-RANSAC} & \bf 73.8 & \bf 39.9 & \bf 97.4 & \bf 0.598 & \bf 3.43\\
        + \ac{USAC} & 70.8 & 35.7 & 97.1 & 0.734 & 3.83\\
        + LM-LO & 72.1 & 37.9 & 97.2 & 0.658 & 3.59 \\
        + \ac{MAGSAC}++ & 66.8 & 30.0 & 96.3 & 0.927 & 4.74\\
\midrule
        \ac{OANet} & 20.7 & \phantom{0}7.6 & 39.7 & 59.0 & 77.9 \\
        + \ac{MAGSAC}++ & 62.6 & 29.0 & 93.2 & 1.06 & 6.99  \\
\bottomrule
\end{tabular}
}
\end{center}
\end{table}

We measure and differentiate performance on indoor and outdoor scenes, which notoriously offer different challenges and exhibit a significant domain gap.

We employ PhotoTourism~\cite{snavely2006photo} as an outdoor dataset, using the same split as in the RANSAC Tutorial from CVPR2020~\cite{cvpr2020ransactutorial}. The dataset consists of several sets of image collections gathered from social media, each set picturing an outdoor landmark. The available ground truth poses are reconstructed and bundle adjusted with COLMAP~\cite{schonberger2016structure}. We apply robust estimation to matches from SuperPoint+SuperGlue~\cite{detone2018superpoint,sarlin2020superglue} and on matches from DoG detections and mutual nearest neighbor matching with HardNet~\cite{mishchuk2017working} descriptors. 
The former scenario presents fewer, more reliable matches, while the latter offers a larger consensus set with a lower inlier rate, and generally better-localized features. The reported numbers are calculated on the 9900 image pairs from the validation split.

In indoor experiments, we use ScanNet~\cite{dai2017scannet}. This dataset consists of 1513 RGB-D scans of indoor environments, densely reconstructed using BundleFusion~\cite{dai2017bundlefusion}. For ScanNet we only use SuperPoint+SuperGlue matches~\cite{detone2018superpoint,sarlin2020superglue}, since we observed the DoG HardNet matches to often fail to provide any signal for model estimation. The reported numbers are calculated on the 1500 image pairs from the test split defined by Sarlin et al.~\cite{sarlin2020superglue}.

\subsection{Ablation Study}
\label{sec:ablations}

We first study the impact of our individual contributions by comparing the following baselines:
%\begin{itemize}
\textbf{Full}: the full proposed pipeline, with the consensus-based state update, inlier-filtered sampling, and inlier-weighted refinement.
\textbf{LM-LO}: a LO-RANSAC~\cite{lebeda2012fixing,chum2003locally} using Levenberg-Marquardt iterative optimization for both inner refinement and final model refinement, and PROSAC sampling. This is the closest traditional baseline to ours.
\textbf{-C}: The full pipeline, without consensus based update. The state are initialized as described, and used to estimate inlier probabilities for sampling and refinement, but they are never updated based on the consensus.
\textbf{-R}: The full pipeline, with a classical refinement weighting strategy instead of our weighting. In this baseline we perform the final refinement with equal weight on all points with Sampson distance smaller than 1 pixel from the initial model.
\textbf{-S}: The full pipeline, with a classical uniform sampler on the correspondences passing a traditional SNN ratio filter of 0.8 instead of our filtering.
%\end{itemize}

\begin{table}[t]
\begin{center}
        \caption{\textbf{Essential matrix} estimation on \textbf{SuperGlue} matches from 1500 images of the \textbf{ScanNet} test set. \ac{CA-RANSAC} (PT) is trained on PhotoTourism to test for generalization.
% We measure pose error as the maximum between rotational and translational error in degrees. We report average error (Avg), median error (Med), Mean Average Precision under 20 degrees (\ac{MAP}@20\degrees), Area under the Curve under 5 degrees (\ac{AUC}@5\degrees) and under 1 degree (\ac{AUC}@1\degrees).
}
\label{tab:scannet_spsg_ess}
\resizebox{\linewidth}{!}{
\begin{tabular}{lccccc}
\toprule
        & AUC5 & AUC1 & MAP20 & Med & Avg \\
\midrule
        \ac{CA-RANSAC} & \bf 19.5 & \bf 0.8 & 77.7 & \bf 6.17 & 18.9\\
        \ac{CA-RANSAC} (PT) & 18.6 & 0.7 & \bf 78.2 & 6.60 & \bf 18.7\\
        \ac{USAC} & 14.0 & 0.4 & 72.5 & 8.32 & 22.0\\
        LM-LO & 14.3 & 0.3 & 72.9 & 8.35 & 21.2\\
        \ac{MAGSAC}++ & 15.4 & 0.6 & 74.0 & 7.82 & 22.7\\
\bottomrule
\end{tabular}
}
\end{center}
\end{table}

\begin{table}[t]
\begin{center}
        \caption{\textbf{Fundamental matrix} estimation on \textbf{SuperGlue} matches from 9900 images of the \textbf{PhotoTourism} validation set. \ac{CA-RANSAC} (E) is trained on Essential matrix to test for generalization.
%We measure pose error as the maximum between rotational and translational error in degrees. We report average error (Avg), median error (Med), Mean Average Precision under 20 degrees (\ac{MAP}@20\degrees), Area under the Curve under 5 degrees (\ac{AUC}@5\degrees) and under 1 degree (\ac{AUC}@1\degrees).
}
\label{tab:photo_spsg_fund}
\resizebox{\linewidth}{!}{
\begin{tabular}{lccccc}
\toprule
        & AUC5 & AUC1 & MAP20 & Med & Avg \\
\midrule
        \ac{CA-RANSAC} & \bf 58.6 & \bf 26.3 & \bf 94.3 & \bf 1.20 & 5.73\\
        \ac{CA-RANSAC} (E) & \bf 58.6 & 25.6 & 94.0 & 1.26 & \bf 5.70\\
        \ac{USAC} & 55.0 & 23.5 & 92.8 & 1.52 & 6.66\\
        LM-LO & 54.3 & 23.2 & 92.9 & 1.55 & 6.60\\
        \ac{MAGSAC}++ & 55.9 & 24.2 & 92.2 & 1.40 & 6.84\\
\bottomrule
\end{tabular}
}
\end{center}
\end{table}

We report the estimation accuracy of these baselines in Table~\ref{tab:ablations}. The \textit{-C} baseline cannot confidently estimate inlier likelihoods without checking for consensus. 
Therefore, the proposed sampling and refinement schemes become detrimental, even after re-training. 
Our proposed sampler in \textit{Full} is clearly superior to the SNN-based sampler of \textit{-S} and to the PROSAC sampler in \textit{LM-LO}. 
The refinement ablation in \textit{-R} shows little difference on DoG detections due to their low localization noise. 
We further compare \textit{-R} with SuperPoint detections on ScanNet in the lower cut of Table~\ref{tab:ablations}. In these conditions, our likelihood-weighted refinement brings very strong improvements on the most precise metrics.

\begin{figure*}
\begin{center}
\includegraphics[width=\linewidth]{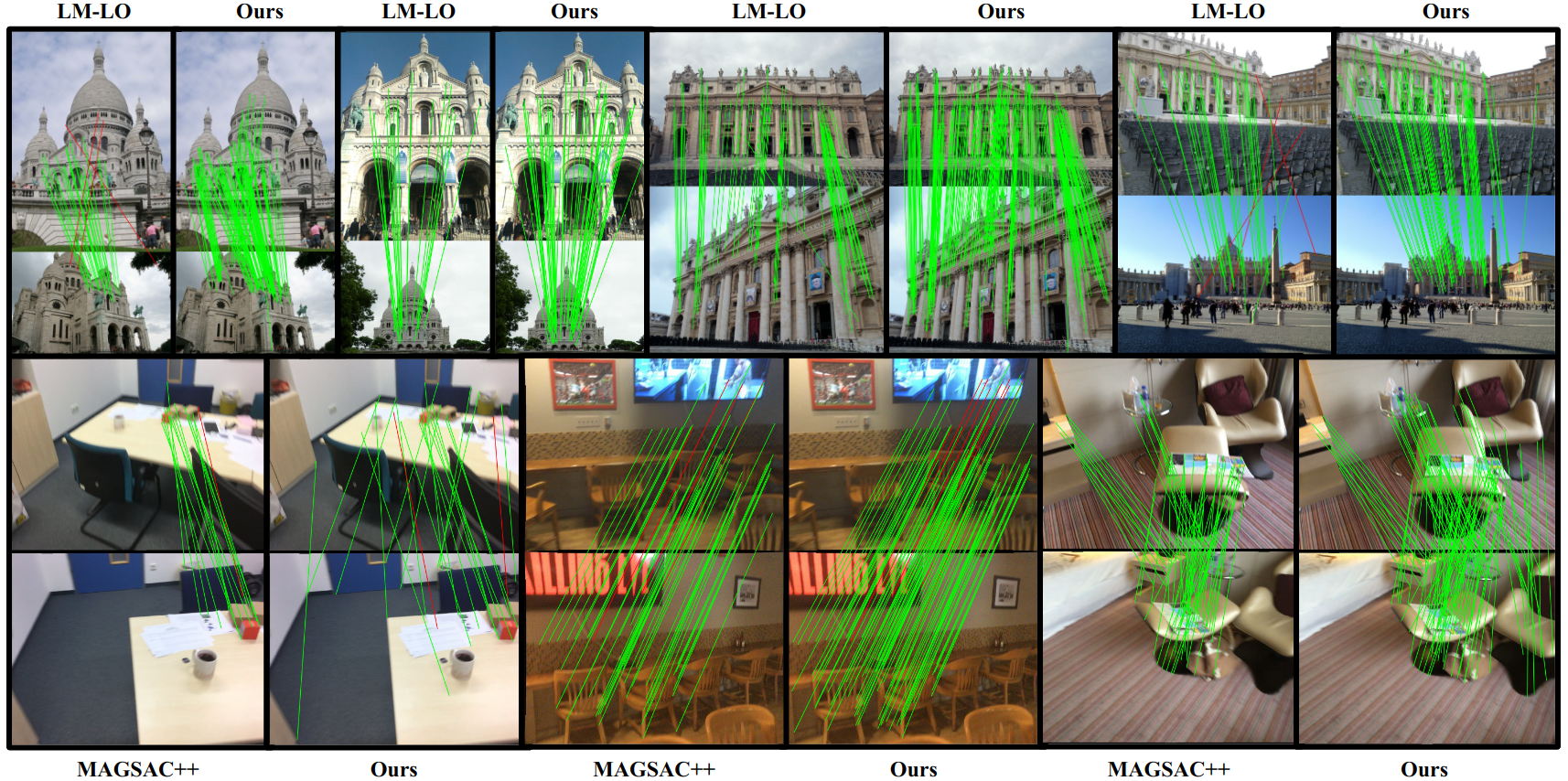}
\end{center}
   \caption{\textbf{Qualitative results}: we compare our method with the strongest alternative baseline on each dataset according to our quantitative evaluation. We draw inlier matches in green and outliers in red. For CA-RANSAC, we only show correspondences with larger than 0.5 estimated likelihood. On PhotoTourism HardNet matches (top row), CA-RANSAC is able to find the full inlier set directly without the aid of an outlier filter, unlike LM-LO which requires pre-filtering most outliers, and therefore misses also many inliers. 
   On ScanNet SuperGlue matches (bottom row) we can generally find a larger consensus set in the same number of iterations.}
\label{fig:qual}
\end{figure*}

\subsection{Comparison with State of the Art}

We compare CA-RANSAC with selected robust estimators to represent the current state of the art. 
USAC~\cite{raguram2013usac} integrates in a single framework progressive sampling~\cite{chum2005matching}, local optimization~\cite{lebeda2012fixing,chum2003locally} and SPRT early termination~\cite{matas2005randomized}. MAGSAC++~\cite{barath2019magsac,barath2019magsacpp} is a recent state-of-the-art RANSAC variant that marginalizes over a range of error thresholds. 
We also include the LM-LO baseline from Section~\ref{sec:ablations}, a variant of LO-RANSAC~\cite{lebeda2012fixing,chum2003locally} which uses progressive sampling and non-linear Levenberg-Marquardt~\cite{more1978levenberg} optimization instead of the sampling-based local optimization traditionally used.
This is the closest non-learned robust estimator to our specific implementation of CA-RANSAC, and we found it to be already very competitive with the current State of the Art, but sensitive to parameter tuning. 
In all baselines, PROSAC~\cite{chum2005matching} sampling is applied on the second nearest neighbor descriptor distance ratio only when direct nearest neighbor matching is applied.

Moreover, we test CLNet~\cite{zhao2021progressive} as a recent learned outlier filter, and show that CA-RANSAC can benefit from tighter integration of a pre-trained CLNet model into its initialization module. 
The original CLNet is trained on RootSIFT matches on PhotoTourism, using the same detector and similar filters as our DoG HardNet mutual nearest neighbors baseline. 
We found that CLNet generalizes to our DoG setting, but we did not observe the same for SuperPoint+SuperGlue matches, so we only report its results on the outdoor experiments with DoG detections.

We perform nonlinear LM~\cite{more1978levenberg} refinement of the output model on the inliers found by each baseline, minimizing the Sampson distance to the model with a Cauchy robust loss. We observed significant improvements on every baseline as a result of this refinement step, so we always perform this final refinement except for Ours and LM-LO, which already include a final LM refinement.

We report experimental results on essential matrix estimation in outdoor conditions with SuperGlue matches in Table~\ref{tab:photo_spsg_ess}. In this setting the inlier rates are relatively high, so the solution refinement has the largest impact. We observe improvements in all monitored metrics, particularly on the less saturated AUC1, showing superior accuracy of our refinement procedure. 

In Table~\ref{tab:photo_dog_ess}, we show results on essential matrix estimation in outdoor conditions with the DoG detector and HardNet mutual nearest neighbor matches. Matches are filtered with the classical SNN filter with threshold 0.85 on all baselines for the best relative pose estimation results, except for CA-RANSAC which learns to use the SNN ratio as side information. In this setting with lower inlier rates we still observe significant improvements in all metrics, with a larger margin on MAP20 due to the more efficient sampling induced by our consensus-aware inlier probability estimates, consistently with the ablations from Table~\ref{tab:ablations}. Note that in this scenario the hard limit on 1024 iterations significantly hinders exploration, however, CA-RANSAC on 1024 iterations is still superior to LM-LO running 100k iterations. As in every other experiment, we disable the termination criterion and force LM-LO to run through all iterations. Some qualitative results from this test are reported in the top row of Figure~\ref{fig:qual}. CA-RANSAC generally finds the largest consensus set from the full set of correspondences, while other baselines suffer from the presence of the SNN filter, which prunes most outliers but also many inliers. Note that the use of such filter is necessary to maximize the estimation performance of every other baseline expect for ours.

Moreover, in Table~\ref{tab:photo_dog_filt_ess} we test on the same setting where a learned outlier filter prunes the great majority of outliers, with better recall than the SNN filter. We select CLNet~\cite{zhao2021progressive} as state of the art outlier filter, which already achieves impressive performance without any robust estimator. With CLNet pruning, we observe strong improvements of every robust estimator. In CA-RANSAC, instead of using CLNet as a filter, we use its pretrained features as side information, to avoid the early decision on pruning. Our method also succeeds to take advantage of the extra signal from CLNet. Similarly to our outdoor experiments with SuperGlue matches, we observe again that in the presence of high inlier rates the largest improvement comes from the refinement stage on low error metrics such as AUC1.

We report experiments on indoor essential matrix estimation in Table~\ref{tab:scannet_spsg_ess} with SuperGlue matches. 
We keep the same parameter and hyperparameter setting as in our outdoor experiments. CA-RANSAC achieves the best robustness and accuracy even when trained on PhotoTourism, and further improves in accuracy metrics after being fine-tuned on the ScanNet training set. Qualitative results on the bottom row of Figure~\ref{fig:qual} show that we can generally find a similar or larger consensus set compared to MAGSAC++, leading to a better conditioning of the final refined model. Note that for clarity in the Figure we only show matches with more than 0.5 confidence to be inliers, but more correspondences also participate in model refinement with lower weight.

We test CA-RANSAC on fundamental matrix estimation in Table~\ref{tab:photo_spsg_fund}. We observe that, simply changing the minimal sample size, the minimal solver, and the refinement constraints to work on fundamental matrices  (labeled as $\text{Ours}^*$), CA-RANSAC seamlessly generalizes from essential matrix without the need to perform any re-training, achieving state-of-the-art results in any metric. Fine-tuning on fundamental matrix estimation helps to improve the estimation accuracy further, however, it is not crucial to get accurate results.

\section{Conclusions}

In this paper, we propose CA-RANSAC, a learned RANSAC variant that benefits from \textit{all} iterations done throughout the RANSAC procedure. 
It conditions its sampling and refinement schemes on rich and adaptively updated state that integrate information about their previous consensus scores. 
We show that our novel formulation of consensus as an attention matrix enables learning effective sampling and refinement schemes with great simplicity. We demonstrate experimentally that CA-RANSAC improves on top of state-of-the-art RANSAC variants, and learns parameters which generalize well to other image domains (outdoor to indoor) and to different models (essential matrix to fundamental matrix).

\begin{figure*}[t]
\begin{center}
\includegraphics[width=\linewidth]{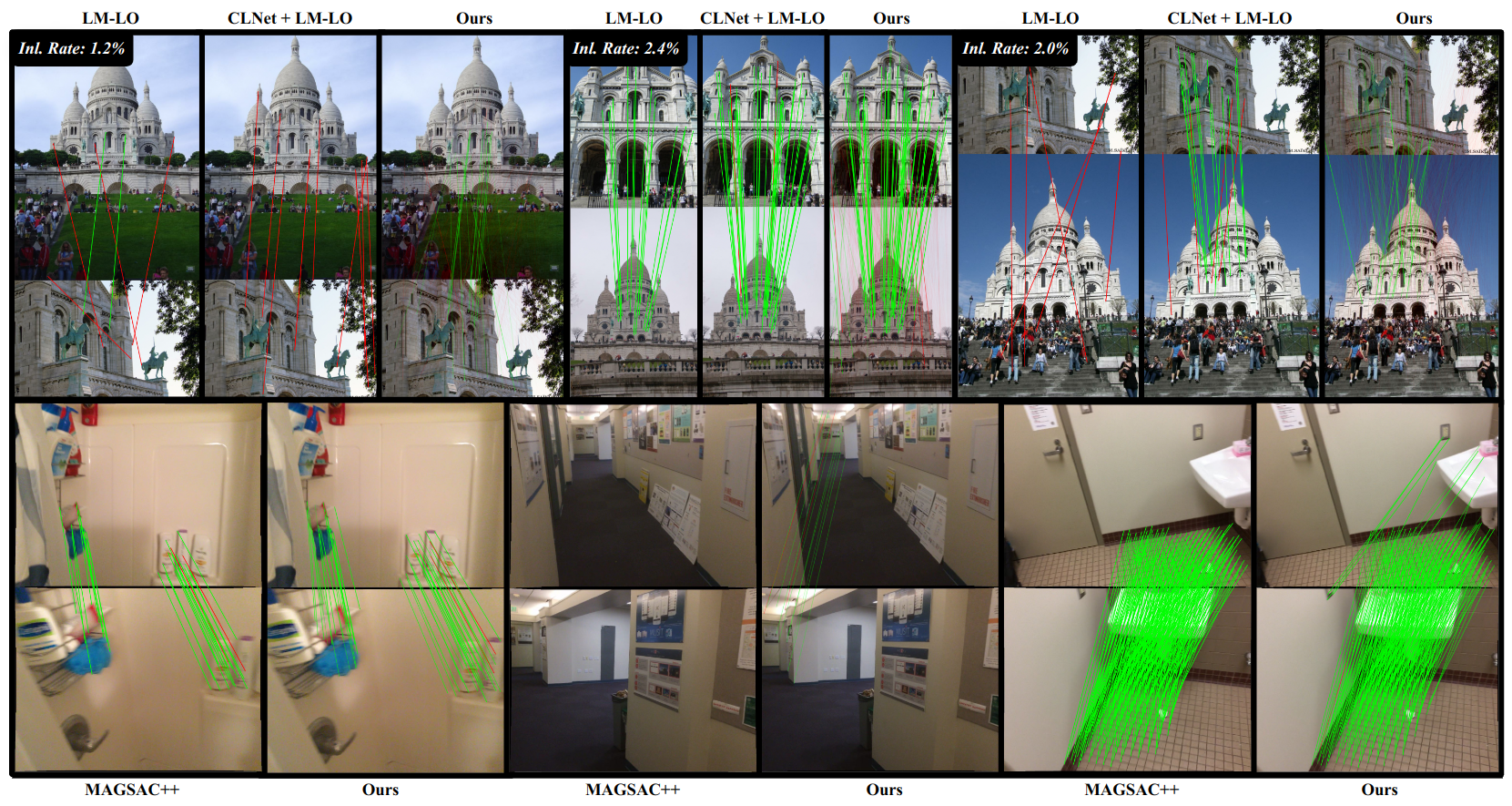}
\end{center}
   \caption{\textbf{Additional qualitative results}: we compare our method with the strongest alternative baseline on each dataset according to our quantitative evaluation. We draw inlier matches in green and outliers in red. For CA-RANSAC, we show all correspondences and encode the estimated inlier likelihood with the line opacity. On PhotoTourism HardNet matches (top row), we mine for examples with extremely low ground truth inlier rates (reported in the top left), and include results from CLNet + LM-LO. The reported CA-RANSAC only uses SNN as side information. We can obtain results comparable to CLNet + LM-LO without any outlier filter, even when the SNN-filtered LM-LO fails. Moreover, the overall likelihood of our matches encodes the confidence of the proposed solution.
On ScanNet SuperGlue matches (bottom row) we propose more comparisons with MAGSAC++ showing the full likelihood-weighted consensus from CA-RANSAC. We find again that the overall likelihood encodes the confidence in the solution for the hardest cases.  }
\label{fig:qual_supp}
\end{figure*}

\newpage

\appendix

\section{Implementation details}
\label{sec:impl_details}

We report here more implementation details and related discussion on top of Section~\ref{subsec:algo}

\paragraph{\textsc{InitState}} The \textsc{InitState} model is implemented as a MultiLayerPerceptron with shared weights for each correspondence. In our implementation, the MLP takes as input a batch of scalars (the SNN ratio for mutual HardNet~\cite{mishchuk2017working} matches, or matching score for SuperGlue~\cite{sarlin2020superglue} matches) which are lifted with fixed Fourier features with exponential frequencies $2^0, 2^1 \dots 2^7$ to improve the MLP sensitivity to small changes. The 16-dimensional Fourier encoding of the side information is then mapped to the initialized 128-dimensional state with a 2-layer MLP with hidden dimension 128 with a single leaky ReLU activation on the one hidden layer. This very simple design is expressive enough to effectively embed a scalar into a curve in the 128-dimensional state space. We experimented with more complex and expressive initialization schemes, and we found that, while they help to improve the results on the first iteration batch, their impact is much lower in successive iterations. Nonetheless, we found that a trivial initialization (e.g. all states are initialized to zero) struggles to find an initial informative consensus when correspondences are numerous and inlier rates are low.

We use this state initialization scheme in all our experiments, with a small variation for Table~\ref{tab:photo_dog_filt_ess} to fuse per-point features from CLNet~\cite{zhao2021progressive} within our initializer. We extract from CLNet the per-correspondence features used to predict the weights for the final weighted 8-point fitting. These 128-dimensional features are available for one fourth of the correspondence, due to the progressive pruning in the CLNet architecture, therefore we pad unavailable features with zeros. Moreover, we take the model residuals estimated from CLNet as further side information, we rescale it and lift with Fourier encoding. We concatenate the CLNet features to the Fourier-lifted side information as input to the same MLP encoder as described previously, where only the input dimensionality is adjusted to fit the extended input.

\paragraph{\textsc{InlierDecoder}} We implement the \textsc{InlierDecoder} as a 3-layer MLP decoding the 128-dimensional state into a scalar in $[0, 1]$, with intermediate hidden dimensionality $64$ and $32$. We use the leaky ReLU activation for hidden units, and a Sigmoid activation on the final output.

\paragraph{\textsc{StateTransformer}} The \textsc{StateTransformer} updates the state $\mathcal{F}$ according to the following structure, also reported in Equation~\eqref{eq:feat_update}:

 \begin{equation} \label{eq:feat_update_numbered}
     \mathcal{F} \leftarrow \text{MLP}_1([\mathcal{F}, \text{MLP}_2(A \cdot \text{MLP}_3(\mathcal{F}))]),
 \end{equation}

 where we number each MLP for easier reference. $\text{MLP}_3$ and $\text{MLP}_2$ have the same 3-layer structure, with constant hidden dimensionality of 128, hyperbolic tangent as hidden activation, and no final activation. $\text{MLP}_1$ uses instead leaky ReLU as hidden activation, and takes as input dimensionality 256 since it fuses the 128-dimensional state with the 128-dimensional attention gated encodings.

 \paragraph{\textsc{Refine}} We perform Levenberg-Marquardt~\cite{more1978levenberg} minimization of the weighted squared Sampson distance of all correspondences with weights $w_i = p_i^\alpha$ where $p_i$ is the estimated inlier probability of correspondence $i$ and $\alpha$ is a single trained parameter initialized to $1.0$. For robustness, we use a Cauchy robust loss to further reduce the influence of outlier points. We observed that more expressive refinement schemes (such as making direct inference of refinement weights with an MLP) do not provide improvements to this baseline, suggesting that the useful signal from consensus for refinement is essentially in terms of inlier probability only. Note that we do not propagate gradients through the term $p_i$ in refinement to ensure that the \textsc{InlierDecoder} learns calibrated inlier probabilities without any biasing signal, and only propagate gradients to $\alpha$ to map $p_i$ to weights. In practice, we run a sparse LM refinement only on points with $w_i > 10^{-3}$ and propagate gradients only through the last refinement iteration.

 Moreover, inspired by LO-RANSAC~\cite{chum2003locally,lebeda2012fixing}, we found that adding an extra refinement before the \textsc{ConsensusAttention} significantly improves results. For efficiency, this intermediate refinement is applied only on the $k$ models with the highest total consensus $C_i$ within the batch, and minimizes the truncated squared Sampson distance of the inliers of each model for a maximum of 10 LM iterations. Then the score matrix $S$ is computed again only on these $k$ models, which is used to compute the attention matrix $A$. We found experimentally that a value as low as $k=4$ is sufficient to grant accurate results, while minimizing the runtime impact of the extra refinement step and attention matrix calculation. 

\section{Additional qualitative results}

We report additional qualitative results in Figure~\ref{fig:qual_supp}. We compare our method with the strongest alternative baseline on each dataset according to our quantitative evaluation in Section~\ref{sec:experiments}. We draw inlier matches in green and outliers in red. For CA-RANSAC, we show all correspondences and encode the estimated inlier likelihood with the line opacity. On PhotoTourism HardNet matches (top row), we mine for examples with extremely low ground truth inlier rates (reported in the top left), and include results from CLNet + LM-LO. The reported CA-RANSAC only uses SNN as side information. We can obtain results comparable to CLNet + LM-LO without any outlier filter, even when the SNN-filtered LM-LO fails. Moreover, the overall likelihood of our matches encodes the confidence of the proposed solution.
On ScanNet SuperGlue matches (bottom row) we propose more comparisons with MAGSAC++ showing the full likelihood-weighted consensus from CA-RANSAC. We find again that the overall likelihood encodes the confidence in the solution for the hardest cases.  

\begin{table*}
\begin{center}
        \caption{\textbf{Essential matrix} estimation on \textbf{HardNet} matches from 9900 images of the \textbf{PhotoTourism} validation set. LM-LO and LM-LO-U uses the \ac{SNN} filter for the best performance, \ac{CA-RANSAC} instead learns to make use of the \ac{SNN} ratio as side information. Each model runs for exactly the specified number of iterations, where X\texttimes Y(\texttimes Z) means that we run X batches of Y iterations each, augmented by Z times through \ac{NeFSAC}. We train \ac{CA-RANSAC} with 4\texttimes 256 iterations and use the same model for every other entry. LM-LO uses \ac{PROSAC} sampling, LM-LO-U uses a uniform sampler.}
\label{tab:photo_dog_ess_iter}
\resizebox{\linewidth}{!}{
\begin{tabular}{lcccccc}
\toprule
        & iterations & AUC5 & AUC1 & MAP20 & Med & Avg \\
\midrule
        \ac{CA-RANSAC} & 1\texttimes 256 & 62.2 & 32.2 & 94.1 & 0.985 & 6.65\\
        \ac{CA-RANSAC} & 2\texttimes 256 & 66.5 & 35.0 & 95.5 & 0.795 & 5.58\\
        \ac{CA-RANSAC} & 4\texttimes 256 & 69.1 & 37.4 & 96.3 & 0.691 & 4.75\\
        \ac{CA-RANSAC} & 8\texttimes 256 & 70.0 & 37.8 & 96.5 & 0.677 & 4.75\\
        \ac{CA-RANSAC} + NeFSAC & 4\texttimes 256(\texttimes 8) & 71.2 & 39.3 & 96.7 & 0.623 & 4.25 \\
        \midrule
        \ac{CA-RANSAC} & 1\texttimes 128 & 58.7 & 30.4 & 92.8 & 1.162 & 7.82\\
        \ac{CA-RANSAC} & 2\texttimes 128 & 64.7 & 34.0 & 94.9 & 0.863 & 6.07\\
        \ac{CA-RANSAC} & 4\texttimes 128 & 67.5 & 35.9 & 95.6 & 0.742 & 5.29\\
        \ac{CA-RANSAC} & 8\texttimes 128 & 69.1 & 36.7 & 95.8 & 0.699 & 5.01\\
        \ac{CA-RANSAC} & 16\texttimes 128 & 69.8 & 36.9 & 95.8 & 0.685 & 4.88\\
        \midrule
        LM-LO & 256 & 63.2 & 33.4 & 91.4 & 0.885 & 8.91\\
        LM-LO & 512 & 64.3 & 34.1 & 91.9 & 0.842 & 8.42\\
        LM-LO & 1024 & 65.0 & 34.7 & 92.4 & 0.808 & 8.20\\
        LM-LO & 2048 & 65.5 & 35.1 & 92.8 & 0.784 & 7.72\\
        LM-LO & 10240 & 66.4 & 35.6 & 93.2 & 0.764 & 7.41\\
        LM-LO & 102400 & 66.6 & 35.7 & 93.6 & 0.761 & 7.06\\
        \midrule
        LM-LO-U & 256 & 60.7 & 32.2 & 89.5 & 0.968 & 10.6\\
        LM-LO-U & 512 & 62.5 & 33.2 & 90.9 & 0.898 & 9.48\\
        LM-LO-U & 1024 & 63.9 & 34.1 & 91.8 & 0.842 & 8.84\\
        LM-LO-U & 2048 & 64.8 & 34.6 & 91.8 & 0.837 & 8.08\\
        LM-LO-U & 10240 & 66.1 & 35.3 & 93.1 & 0.777 & 7.35\\
        LM-LO-U & 102400 & 66.6 & 35.6 & 93.5 & 0.768 & 6.99\\
\bottomrule
\end{tabular}
}
\end{center}
\end{table*}

\begin{table*}
\begin{center}
        \caption{\textbf{Essential matrix} estimation on weaker matches from 9900 images of the \textbf{PhotoTourism} validation set.
        We measure pose error as the maximum between rotational and translational error in degrees. We report average error (Avg), median error (Med), Mean Average Precision under 20 degrees (\ac{MAP}@20\degrees), Area under the Curve under 5 degrees (\ac{AUC}@5\degrees) and under 1 degree (\ac{AUC}@1\degrees). Every baseline uses a further \ac{SNN} ratio filter of 0.85, except for \ac{CA-RANSAC} which uses only the \ac{SNN} as side information.
}
\label{tab:photo_weak_kp}
\resizebox{\linewidth}{!}{
\begin{tabular}{lcccccc}
\toprule
        & keypoints & AUC5 & AUC1 & MAP20 & Med & Avg \\
        \midrule
        \ac{MAGSAC}++ & RootSIFT 2k & 56.2 & 27.0 & 88.4 & 1.29 & 11.3\\
        USAC & RootSIFT 2k & 59.6 & 30.7 & 90.1 & 1.09 & 10.0\\
        VSAC & RootSIFT 2k & 58.7 & 30.2 & 91.3 & 1.12 & \phantom{0}9.3\\
        GC-RANSAC & RootSIFT 2k & 62.5 & 33.0 & 91.5 & 0.90 & \phantom{0}8.8\\
        LM-LO & RootSIFT 2k & 65.1 & \bf 34.8 & 92.9 & \bf 0.80 & \phantom{0}8.0\\
        \ac{CA-RANSAC} & RootSIFT 2k & \bf 65.8 & 34.7 & \bf 94.8 & 0.81 & \bf \phantom{0}6.5\\
        \midrule
        \ac{MAGSAC}++ & \acs{ORB} 2k & \phantom{0}8.5 & 2.1 & 32.0 & 53.6 & 69.5\\
        USAC & \acs{ORB} 2k & \phantom{0}8.3 & 2.0 & 33.8 & 47.5 & 67.7\\
        VSAC & \acs{ORB} 2k & \phantom{0}9.6 & 2.5 & 36.3 & 42.4 & 66.4\\
        GC-RANSAC & \acs{ORB} 2k & 12.8 & 4.0 & 38.0 & \bf 41.4 & 65.8\\
        LM-LO & \acs{ORB} 2k & 14.2 & 4.4 & 38.3 & 45.2 & 67.3\\
        \ac{CA-RANSAC} & \acs{ORB} 2k & \bf 14.9 & \bf 4.5 & \bf 39.6 & \bf 41.4 & \bf 65.3\\
        \midrule
        \ac{MAGSAC}++ & \acs{ORB} 8k & 18.1 & \phantom{0}4.8 & 55.2 & 14.3 & 42.2\\
        USAC & \acs{ORB} 8k & 16.7 & \phantom{0}4.2 & 54.8 & 15.4 & 42.9\\
        VSAC & \acs{ORB} 8k & 19.2 & \phantom{0}5.2 & 58.9 & 12.7 & 39.9\\
        GC-RANSAC & \acs{ORB} 8k & 29.6 & 11.3 & 62.1 & \phantom{0}8.1 & 38.7\\
        LM-LO & \acs{ORB} 8k & 31.1 & 11.2 & 52.9 & \phantom{0}6.3 & 33.4\\
        \ac{CA-RANSAC} & \acs{ORB} 8k & \bf 33.3 & \bf 12.3 & \bf 69.5 & \bf \phantom{0}5.3 & \bf 31.1\\
\bottomrule
\end{tabular}
}
\end{center}
\end{table*}

\section{Processing Time}
\label{sec:processing_time}

\begin{figure}[t]
\begin{center}
   \includegraphics[width=0.9\linewidth]{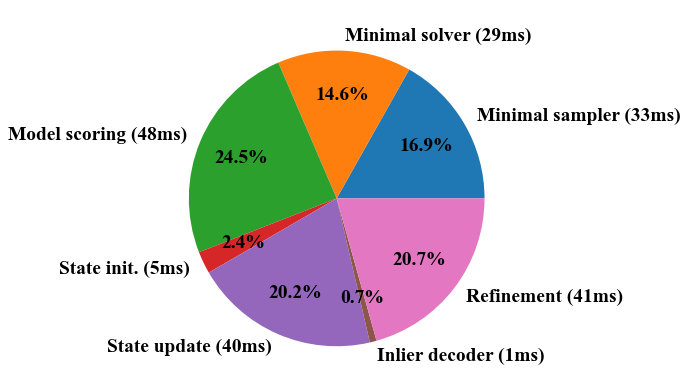}
\end{center}
   \caption{Runtime breakdown of each component in CA-RANSAC. Times are averaged over 9900 image pairs with $\approx2000$ correspondences per image pair, on a i9-9900K CPU, using our PyTorch implementation for 1024 iterations. The proposed components (state initialization, state update, and inlier decoder) account for less than 25\% of the total estimation runtime of 197ms.}
\label{fig:runtime_pie}
\end{figure}

When adding neural networks within time-critical components, it is important to monitor the impact of the network inference times on the total processing time. 
Since we study the sample efficiency running on a fixed number of iterations, in this paper, we do not consider any early termination criterion which could reduce the total runtime. % of CA-RANSAC.

A major design choice minimizing the runtime of our learned contributions is batching of iterations: we update the state based on the consensus collected from the previous 256 iterations, thus, our lightweight transformer is applied only 4 times in 1024 iterations. 
Moreover, we observed that shallow neural networks and state dimensionality as low as 128 are sufficient to achieve good results.

In Figure~\ref{fig:runtime_pie}, we visualize a breakdown of the processing time for each component of CA-RANSAC on HardNet matches on PhotoTourism. The runtimes are averaged over all test image pairs, always running all 1024 iterations. We use our PyTorch implementation of CA-RANSAC on a i9-9900K CPU, therefore no GPU parallelization nor any compiler optimizations are used.

We observe a direct impact of less than 25\% of the total runtime coming from our contributed learned components. While the total runtime measured on our implementation is still higher compared to well-engineered RANSAC implementations with several algorithmic and compiler optimizations, we demonstrate that it is possible to learn an effective conensus-adaptive behavior in RANSAC without a large relative impact on the total runtime.

\section{Influence of number of iterations} \label{sec:iterstudy}

In this Section, we study the influence of the number of iterations on CA-RANSAC compared to its closest classical baseline LM-LO on mutual HardNet matches on the PhotoTourism dataset.
Note that this is the same setting as Table~\ref{tab:photo_dog_ess}.
We report in Table~\ref{tab:photo_dog_ess_iter} the results for CA-RANSAC and LM-LO running on different iterations budget.
Each model runs for exactly the specified number of iterations, where X\texttimes Y(\texttimes Z) means that we run X batches of Y iterations each, augmented by Z times through NeFSAC.
We train CA-RANSAC with 4\texttimes 256 iterations and use the same model for every other entry.
LM-LO uses PROSAC sampling, LM-LO-U uses a uniform sampler.

We can observe very different patterns in the effect of the number of iterations across different methods.
First, we notice that the batching of iterations in CA-RANSAC does have a significant impact.
In fact, we can observe that running a single batch (1\texttimes 256) is significantly worse than running multiple batches even with the same number of total iterations (2\texttimes 128).
This is due to the fact that the proposed sampler only has the chance to sample from the initialized inlier pool, and never has a chance to sample from a consensus-updated inlier pool when only a single batch of iterations is applied.
On the contrary, breaking abundant iterations in many smaller batches of updates does not bring a consistent advantage compared with less large batches, while bringing a significant computational cost.
Between CA-RANSAC on 2\texttimes 256 iterations and its saturation level at 4\texttimes 256(\texttimes 8) we observe a gap of $4.7$ in AUC@5\degrees, similar to the gap of $4.1$ between LM-LO-U at 512 and 102400 iterations, and much larger than the equivalent gap of $2.3$ in LM-LO.
This analogy suggests that there is room for improving the uniform sampling scheme used in CA-RANSAC towards a more efficient biased sampling scheme even within the inlier pools.
Indeed, we observe that PROSAC sampling greatly improves the efficiency of the early iterations in LM-LO compared to the uniform scheme in LM-LO-U and CA-RANSAC with respect to the possible accuracy at saturation.
Given the ablation in Table~\ref{tab:ablations}, which showed the little influence of the novel refinement scheme on this specific validation setting, we can attribute the gap in the accuracy with abundant iterations to the dynamic pool of inliers accounted within CA-RANSAC, in contrast with the fixed pool given by the SNN ratio filter in LM-LO.

\section{Experimental study on weaker descriptors} \label{sec:weak_desc}

In this Section, we consider the option to work with weaker matchers than in the main Chapter. In particular, we measure and compare the performance of CA-RANSAC in the following settings:
\begin{itemize}
        \item \textbf{RootSIFT 2k}: this is the data from the IMW2020 PhotoTourism challenge. Correspondences are obtained using RootSIFT features and MNN filtered first nearest neighbor matching. The final number of correspondences is approximately 2000. Every baseline uses a further SNN ratio filter of 0.85, except for CA-RANSAC which uses only the SNN as side information.
        \item \textbf{ORB 2k}: we extract exactly 2000 ORB keypoints and descriptors on every image, and perform purely first nearest neighbor matching. This produces exactly 2000 correspondences for every image pair. Every baseline uses a further SNN ratio filter of 0.85, except for CA-RANSAC which uses only the SNN as side information.
        \item \textbf{ORB 8k}: we extract exactly 8000 ORB keypoints and descriptors on every image, and perform purely first nearest neighbor matching. This produces exactly 8000 correspondences for every image pair. Every baseline uses a further SNN ratio filter of 0.85, except for CA-RANSAC which uses only the SNN as side information.
\end{itemize}

We report results in Table~\ref{tab:photo_weak_kp}. We observe LM-LO and GC-RANSAC to be particularly competitive in these scenarios. We find CA-RANSAC to be still marginally superior in almost all metrics to its closest classical baseline LM-LO.

\newpage 

%%%%%%%%% REFERENCES
{\small
\bibliographystyle{ieee_fullname}
\bibliography{egbib}
}

\end{document}